\documentclass[runningheads]{llncs}
\usepackage[T1]{fontenc}
\usepackage{graphicx}
\usepackage{amssymb,amsfonts}
\usepackage{algorithmic}
\usepackage{textcomp}
\usepackage{makecell}
\usepackage[utf8]{inputenc}
\usepackage{tabularray}
\usepackage{enumitem}
\usepackage{caption, subcaption}
\usepackage{booktabs, multirow, multicol}
\usepackage{url}

\usepackage[pagebackref,breaklinks,colorlinks]{hyperref}
\usepackage{xcolor}

\hypersetup{citecolor=black}
\hypersetup{
    colorlinks=true,
    linkcolor=black,
    urlcolor=black,
}

\begin{document}

\title{gWaveNet: Classification of Gravity Waves from Noisy Satellite Data using Custom Kernel Integrated Deep Learning Method}

\titlerunning{gWaveNet: Gravity Wave Classification using Custom Kernel}

\author{Seraj Al Mahmud Mostafa\inst{1} \and
Omar Faruque\inst{1} \and
Chenxi Wang\inst{2} \and
Jia Yue\inst{3,4} \and
Sanjay Purushotham\inst{1} \and
Jianwu Wang\inst{1}}

\authorrunning{SAM Mostafa et al.}

\institute{Department of Information Systems, University of Maryland, Baltimore County, Baltimore, MD, USA \and
Goddard Earth Sciences Technology and Research (GESTAR) II, University of Maryland, Baltimore County, Baltimore, MD, USA \and
Department of Physics, Catholic University of America, Washington, DC, USA \and
NASA Goddard Space Flight Center, Greenbelt, MD, USA\\
\email{\{serajmostafa, omarfaruque, chenxi, psanjay, jianwu\}@umbc.edu, jia.yue@nasa.gov}}

\maketitle
%
\begin{abstract}
Atmospheric gravity waves occur in the Earth's atmosphere caused by an interplay between gravity and buoyancy forces. These waves have profound impacts on various aspects of the atmosphere, including the patterns of precipitation, cloud formation, ozone distribution, aerosols, and pollutant dispersion. Therefore, understanding gravity waves is essential to comprehend and monitor changes in a wide range of atmospheric behaviors. Limited studies have been conducted to identify gravity waves from satellite data using machine learning techniques. Particularly, without applying noise removal techniques, it remains an underexplored area of research. This study presents a novel kernel design aimed at identifying gravity waves within satellite images. The proposed kernel is seamlessly integrated into a deep convolutional neural network, denoted as \textbf{gWaveNet}. Our proposed model exhibits impressive proficiency in detecting images containing gravity waves from noisy satellite data without any feature engineering. The empirical results show our model outperforms related approaches by achieving over 98\% training accuracy and over 94\% test accuracy which is known to be the best result for gravity waves detection up to the time of this work. We open sourced our code at {https://rb.gy/qn68ku}.

\end{abstract}

\section{Introduction}
\label{sec:intro}

Gravity waves (GW) are physical perturbations caused by gravity's restoring force in a planetary environment, distinct from gravitational waves \cite{jovanovic2018nature}. In Earth's atmosphere, various disturbances like airflow over mountains, jet streams, and thunderstorms create atmospheric gravity waves, displacing air parcels and leading to wave patterns resembling ripples on water \cite{mann2019}. These waves have broad effects, including localized vertical motion, turbulence, and impacts on the middle atmosphere's dynamics. They contribute to the transport of heat, momentum, and atmospheric composition \cite{alexander1997model}, as well as influencing weather patterns, precipitation, cloud formation, tidal waves \cite{Fritts2003}, and aviation safety due to clear air turbulence \cite{Moran2018}. Due to the significant impact of gravity waves, there has been a surge of interest in their detection. AI researchers, along with domain experts, are using machine learning techniques to understand the phenomenon better and improve detection accuracy. However, the single-channel satellite dataset used in this study presents challenges. Firstly, limited ground truth availability makes data accuracy verification difficult. Secondly, the dataset contains significant noise interference such as city lights, clouds, and instrumental horizontal/vertical lines. Lastly, there's a restricted amount of data identified by domain experts. While gravity wave data are publicly accessible \cite{data_src}, the ground truth is not provided. Domain experts helped select data containing gravity wave patterns. The dataset comprises night bands obtained from the VIIRS satellite's day/night band (DNB) \cite{gravity_wave_data}, introducing noise from city lights and clouds that may reduce classification accuracy \cite{hasan2022noise}. Applying denoising methods, like Fast Fourier Transform (FFT) \cite{gonzalez2022atmospheric}, can blend gravity wave patterns with noise, making them harder to isolate.

To enhance the classification of gravity waves in satellite images, we introduce a specialized convolutional kernel, the checkerboard kernel. This custom kernel is designed to improve pattern recognition during convolution, particularly for complex features and noisy environments. Inspired by successful applications in computer vision, such as depth completion and image classification, our approach emphasizes the importance of tailored kernels. Studies by Ku et al. \cite{CC-DBLP2018}, Pinto et al. \cite{pinto2011comparing}, and Zhang et al. \cite{CC-zhang20113d} underscore the effectiveness of custom kernels in diverse scenarios. In our specific application, the use of the custom kernel enhances classification accuracy for gravity waves in satellite images, capturing finer details and outperforming conventional methods. While deep neural networks (DNNs) excel in various tasks  \cite{tushar2024cloudunet}, \cite{mostafa2024yolobasedoceaneddy}, \cite{li2022enhanced}, \cite{khan2023flood}, but may not always be optimal for detecting shapes in images due to the extensive training data required. Effective categorization and minimizing assumptions are crucial, as advised by \cite{o2020deep}. In situations with limited data or subtle shapes, deep learning may struggle, suggesting the need for complementary techniques \cite{marcus2018deep}. Additionally, adjusting weights to amplify signals enhances learning \cite{zohuri2020deep}. In our study with noisy data, we propose a hybrid method using a custom kernel to capture challenging shape information for deep neural networks. We focus on accurately identifying gravity waves within images, even in noisy conditions. We introduce `gWaveNet', a hybrid deep neural network integrating the `checkerboard' kernel in the first layer. Our main contributions are: \textit{1) We designed a unique `checkerboard' kernel capable of detecting gravity wave features amidst noise.} This kernel acts as a specialized filter, highlighting gravity wave patterns for more accurate detection. \textit{2) We propose }`gWaveNet'\textit{, a novel deep neural network incorporating our custom-designed checkerboard kernel to enhance gravity wave detection.} This integration allows the model to effectively learn and recognize intricate wave patterns. \textit{3) We conducted extensive ablation studies, exploring various training configurations (as discussed in Section~\ref{resutls}), employing checkerboard kernels of different sizes to demonstrate our model's performance.} These experiments helped us understand the impact of different setups and modifications needed to address the challenges of detecting gravity waves in noisy datasets. Traditionally, detecting gravity waves required expert knowledge and handcrafted features tailored to wave characteristics \cite{lee2017deep}. However, our model automatically learns and extracts relevant features from the data, reducing the need for domain-specific knowledge and enhancing our approach's generalizability.

The rest of the paper is structured as follows. Section \ref{sec:related-works} reviews relevant literature related to this research. Section \ref{sec:dataprep} discusses the facts about data, its collection process, preprocessing steps, and groundtruth information. In Section \ref{methods}, we detail the methodologies employed for both the proposed kernel and the model. Experiment details and results discussed in Section \ref{resutls}. Lastly, we conclude the paper in Section \ref{conclusions}.

\section{Related Works}
\label{sec:related-works}
In this section, we review related work that is particularly relevant to custom kernels and the detection of gravity waves.

\textbf{Custom Kernel for Image Detection.}
Yousafzai et al. proposed a polynomial custom kernel based on Mercer’s theorem with the support vector machine for acoustic waveform classification with noise signals \cite{CC-yousafzai2009custom}. Ku et al. proposed a simple algorithm to generate the depth information of LIDAR sensor data and outperformed deep learning-based methods \cite{CC-DBLP2018}, which applied four $(5\times5)$ custom kernels of circle, cross, diamond, and full shapes to compute the missing data points in sensor data which improved the accuracy of depth information. Suresha et al. proposed the integration of a custom multiquadric kernel function ($k=\sqrt{\left \| x_1-x_j \right \|^{2}+c^2}$) with the KPCA algorithm to generate important features for image classification \cite{CC-suresha2022mq}. This custom kernel which resembles the sigmoid kernel helps to extract new features from the original feature space and also increases classification accuracy and reduces computational complexity, time complexity, and storage issues. Zhang et al. \cite{CC-zhang2007local} performed texture and object classification using kernel-based discriminative analysis of local features to make the distinction between different classes. Also, Zhang et al. \cite{CC-zhang20113d} proposed the integration of custom kernels with the 3D depth information to better analyze and classify 3D objects in the presence of noise and low-intensity data.

\textbf{Deep Learning Approaches for GW Detection.} 
CNN models often improve object detection performance greatly in the computer vision domain. As gravity waves can be detected by analyzing satellite images, the CNN model can be used for this task very efficiently. Thus far, some research has been conducted regarding gravity wave detection using deep learning models. There is a notable study by Lai et al. that developed a convolutional neural network-based auto-extraction program, which extracts gravity wave patterns in all-sky airglow images~\cite{lai2019automatic}. In this work, the process involves using cleaner images, and on top of that, there is still a step of discarding images that are not suitable for the model to learn during the training process. In our case, we fed all images that include noise. Matsuoka et al. used U-net deep neural network model to estimate the gravity wave from reanalysis data \cite{matsuoka2020ARA}. Recently, paper~\cite{sreekanth2023DicL} proposed Deep Dictionary Learning algorithm to integrate deep learning and dictionary learning to detect gravity waves by learning different kernels. InceptionV3 deep learning model was used in \cite{gonzalez2022atmospheric} with transfer learning technique to detect gravity waves from satellite images. In addition to these advancements, attention-based techniques, particularly the Transformer, have gained popularity for image classification. Dosovitskiy et al. \cite{dosovitskiy2020image} demonstrated that Transformers can outperform CNN models. Chen et al. \cite{chen2021crossvit} highlighted the potential of Transformers in capturing multiscale features from images. In the context of remote sensing, Bazi et al. \cite{bazi2021vision} introduced a Transformer-based classification model. In our research, we also evaluated the performance of the Vision Transformer on complex satellite dataset to assess its suitability.

In conclusion, both custom kernel-based methods and deep learning-based approaches have their strengths and weaknesses in image classification tasks. Custom kernels can capture fine-grained details in the images and are especially useful when the shape information is subtle. On the other hand, deep learning-based approaches are highly flexible and can learn complex representations of the images given enough training data. Our study seeks to integrate both techniques to perform gravity wave detection from noisy satellite data with higher accuracy. 

\section{Data Preprocessing}
\label{sec:dataprep}

\begin{figure}
    \begin{minipage}[t]{0.24\columnwidth}
      \includegraphics[width=\linewidth, height=3cm]{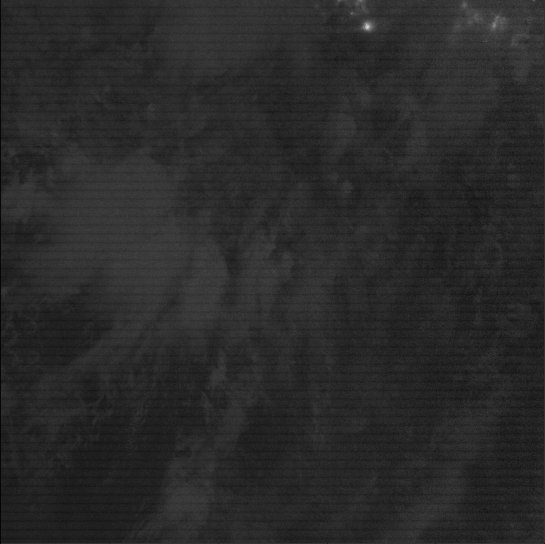}
      \subcaption{normalized}
      \label{subfig:norm}
    \end{minipage}\hfill 
    \begin{minipage}[t]{0.24\columnwidth}
      \includegraphics[width=\linewidth, height=3cm]{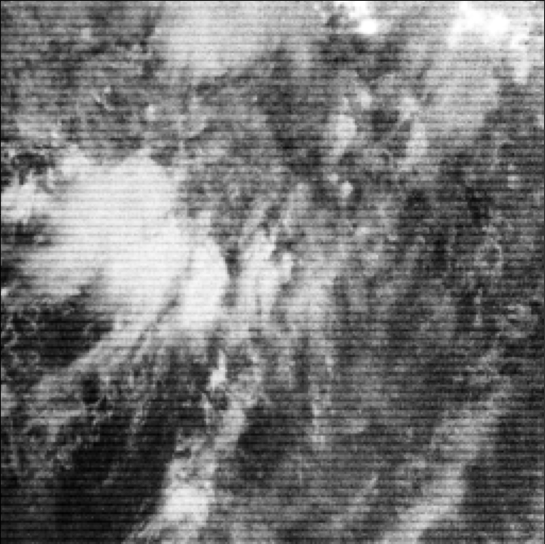}
      \subcaption{pre-processed}
      \label{subfig:prep}
    \end{minipage}
    \begin{minipage}[t]{0.24\columnwidth}
      \includegraphics[width=\linewidth, height=3cm]{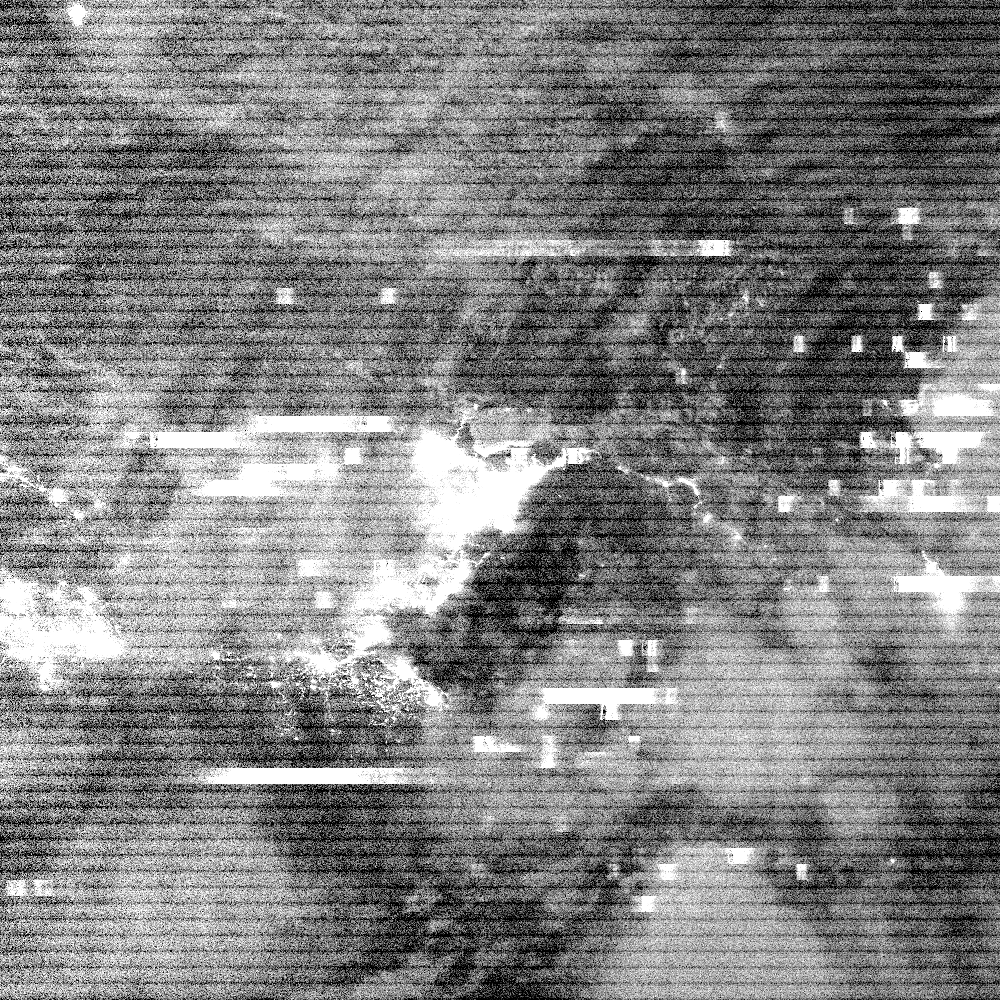}
      \subcaption{gravity waves}
      \label{subfig:gw}
    \end{minipage}\hfill 
    \begin{minipage}[t]{0.24\columnwidth}
      \includegraphics[width=\linewidth, height=3cm]{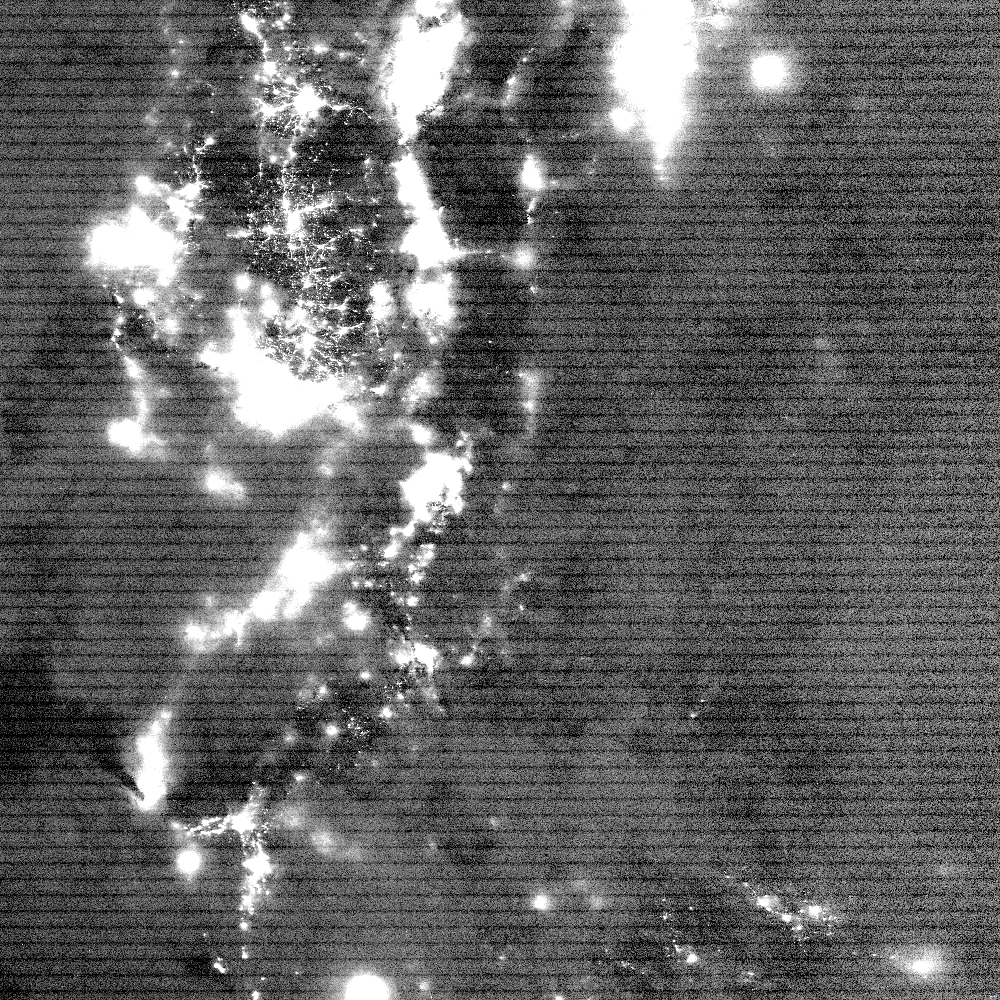}
      \subcaption{no gravity waves}
      \label{subfig:ngw}
    \end{minipage}
    \caption{Examples of Data Processing and Image Types:
(a) Normalized data from raw HDF5 file
(b) Pre-processed image from the same file
(c) Image with gravity waves, including unwanted elements (clouds, city lights, instrumental noise)
(d) Image without gravity waves.}
    \label{fig:prep-n-sample}
    \vspace*{-.5cm} 
\end{figure}

For this investigation, we used the Day/Night Band (DNB) images from the Visible Infrared Imaging Radiometer Suite (VIIRS) instrument onboard the Suomi NPP satellite \cite{gravity_wave_data}. VIIRS DNB observes board band upwelling radiance in the visible region. VIIRS DNB has a wide swath ($\sim$3,000 km) and a relatively high spatial resolution at 1 km approximately. Pixels within a 6-minute granule ($\sim$4,000 x 3,000 pixels) are stored in one Hierarchical Data Format version-5 (HDF5) \cite{hdf5} file. The raw HDF5 files contain radiance measurements within the wavelength range of $0.5\mu m$ to $0.9\mu m$. To highlight the airglow from gravity wave events, nighttime images under new moon conditions are used in this study. As a result, the DNB radiance could be extremely low with a value in the order of magnitude of $-10^{-9}$W/$cm^{-2}$$sr^{-1}$. To comprehend easily, we performed preprocessing on the raw data, ensuring that the array values are within a specific range while maintaining their relative distribution. This involved subtracting the minimum value from all array elements, scaling by the median, and normalizing to 0.5. Normalizing to this reference point enables easier visual comparison and analysis of the data. Finally, we transformed the intensity distribution from an approximate normal distribution to a uniform one, while preserving the accurate range of values. We present examples of our data processing in Figure \ref{fig:prep-n-sample}. Sub-figure \ref{subfig:norm} shows normalized data from a raw HDF5 file, while sub-figure \ref{subfig:prep} displays a pre-processed image derived from the same file. Sub-figure \ref{subfig:gw} illustrates an image containing gravity waves, along with various unwanted elements such as clouds, city lights, and instrumental noise. Finally, sub-figure \ref{subfig:ngw} presents an image without gravity waves. The algorithm for raw data preprocessing is detailed in our previous work \cite{gonzalez2022atmospheric}.

We started by gathering raw satellite data stored in HDF5 format, focusing on 50 files chosen by domain experts containing gravity waves and noise. Using the GDAL library we normalize the HDF5 files and convert them into PNG format. We generated 200x200 grayscale image patches, overcoming the limited dataset challenge. Given the infrequent occurrence of gravity waves, we employed data augmentation, including rotation and flip, to increase the number of patches with gravity waves. We manually categorized patches into two classes: "gw" for gravity waves and "ngw" for non-gravity waves, maintaining a balanced dataset of 5,985 image patches in each class, resulting 11,970 in total.

\section{Research Methodology}
\label{methods}

\begin{figure}
    \centering
    \begin{minipage}[b]{0.36\linewidth}
        \centering
        \includegraphics[width=\textwidth]{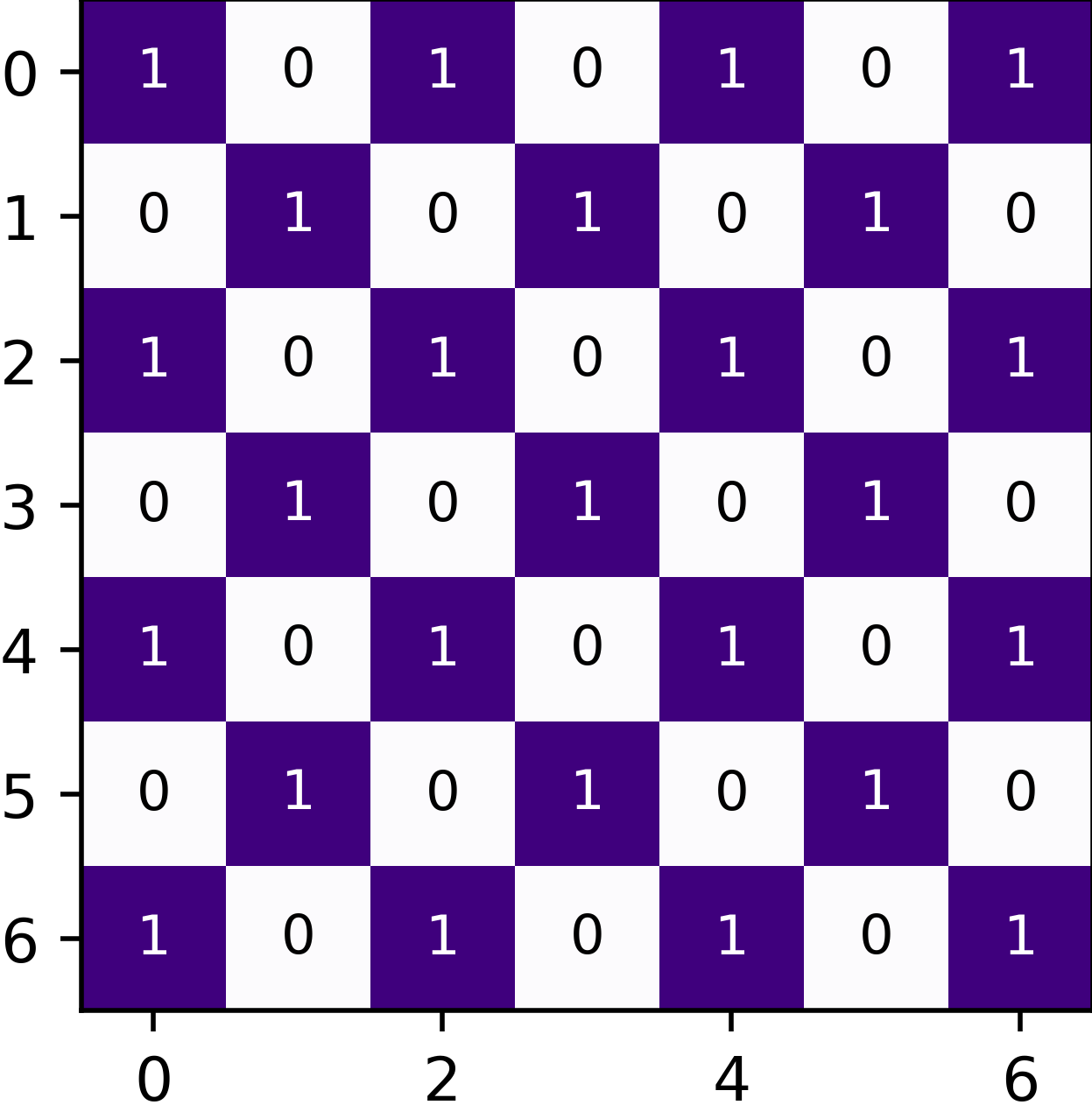}
        \subcaption{7x7 kernel}
    \end{minipage}
     \hspace{0.02\textwidth}
    \begin{minipage}[b]{0.36\linewidth}
        \centering
        \includegraphics[width=\textwidth]{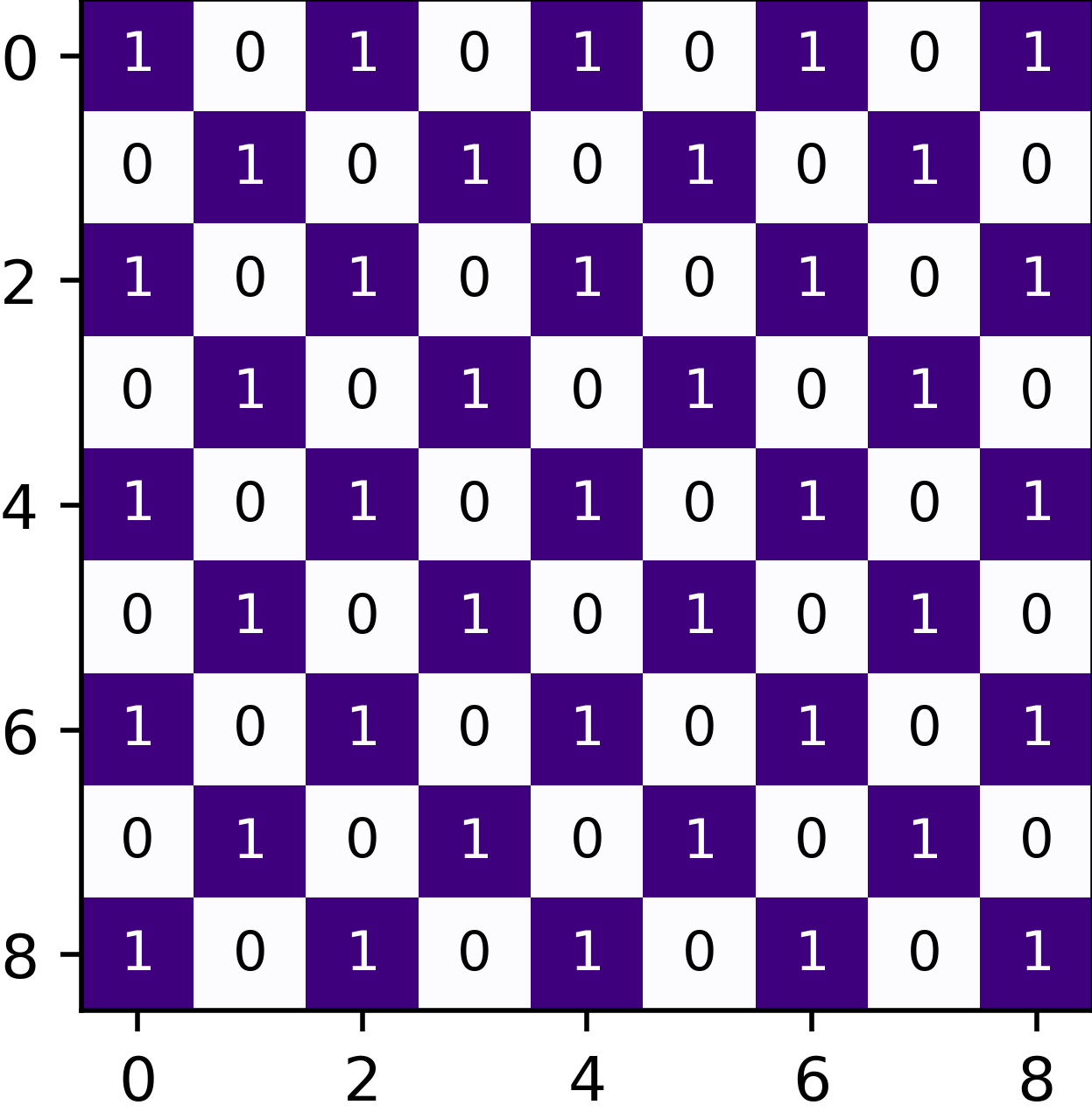}
        \subcaption{9x9 kernel}\par
    \end{minipage}
    \caption{Examples of `checkerboard' kernels proposed in the \textit{gWaveNet}.}
  \label{fig:ck}
\vspace*{-.5cm} 
\end{figure}
\vspace{-.2cm} 

\begin{figure}
    \centering
    \begin{minipage}[b]{0.36\linewidth}
        \centering
        \includegraphics[width=\textwidth]{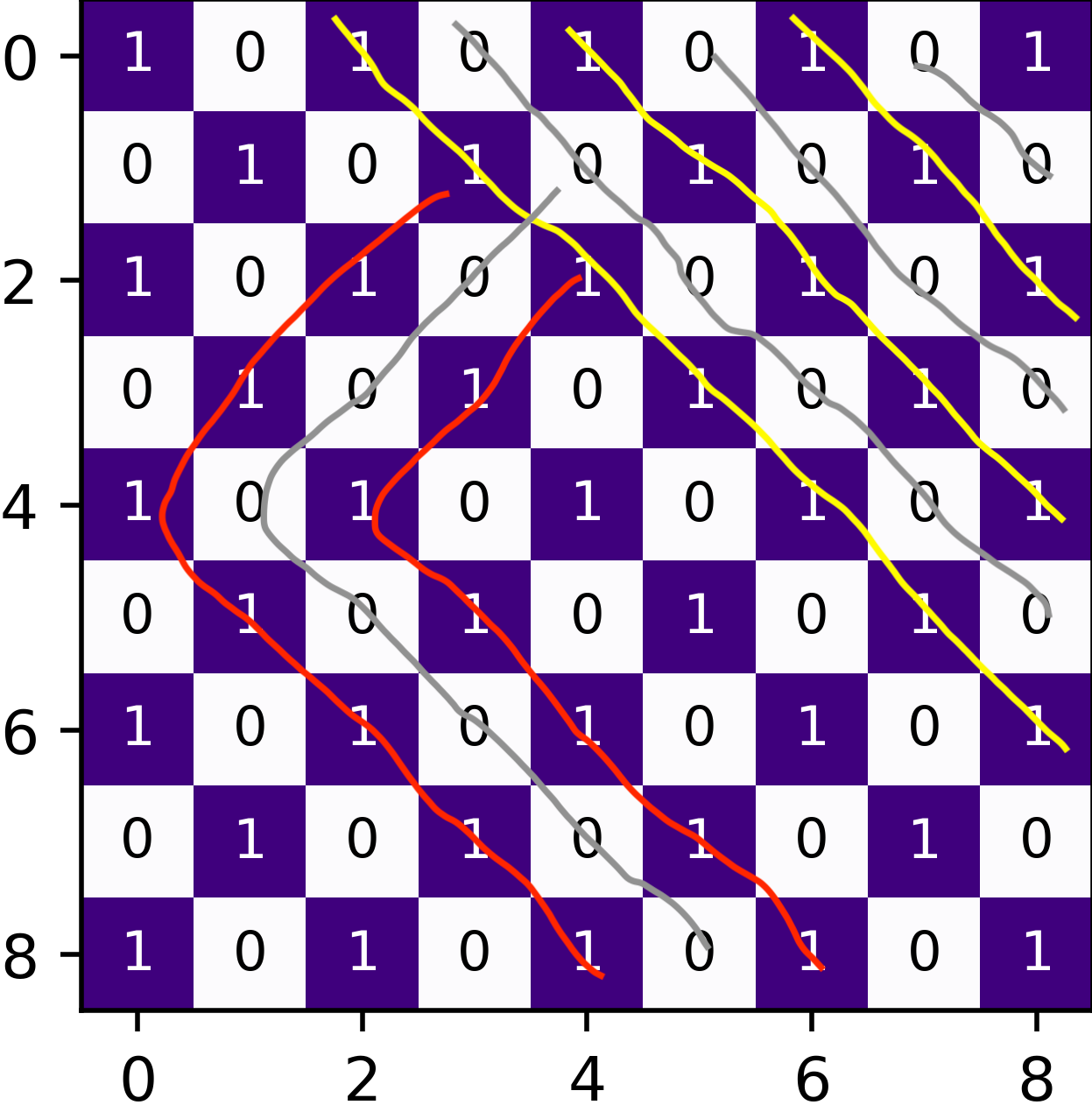}
          \subcaption{}
  \label{fig:illust}
    \end{minipage}
     \hspace{0.02\textwidth}
    \begin{minipage}[b]{0.44\linewidth}
        \centering
        \includegraphics[width=\textwidth]{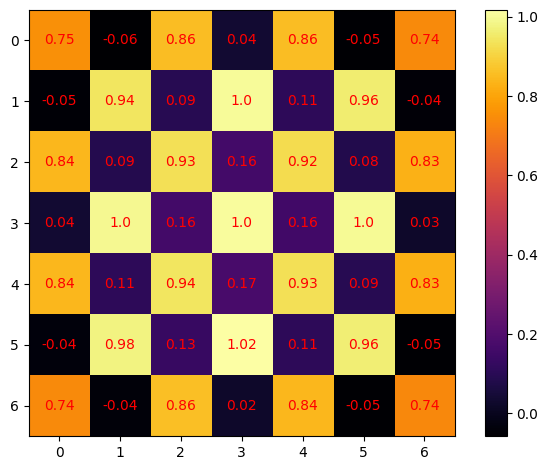}
          \subcaption{}
  \label{fig:trained_kernel}\par
    \end{minipage}
    \caption{Checkerboard kernel concept and application. (a) Proposed kernel capturing gravity wave patterns: yellow lines - potential linear waves, red lines - potential non-linear waves, gray lines - wave gaps. (b) Learned kernel from trained \textit{gWaveNet}.}
  \label{fig:ckexpectations}
  \vspace*{-.75cm} 
\end{figure}

\begin{figure*}[h]
  \begin{minipage}{.22\linewidth}
    \includegraphics[width=3.5cm, height=4cm]{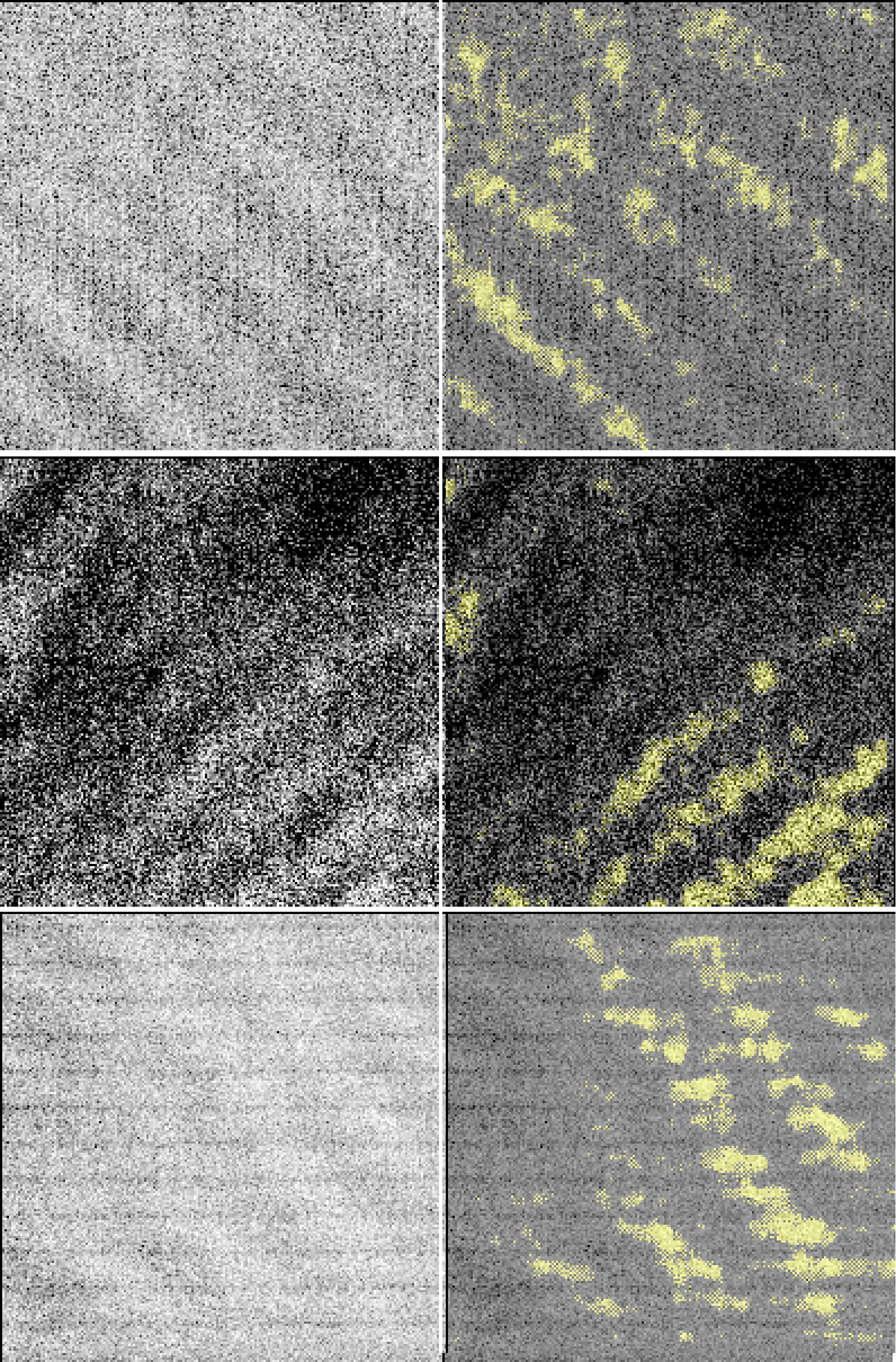}%
    \subcaption{easy detect}%
  \end{minipage}\hfil
  \begin{minipage}{.22\linewidth}
    \includegraphics[width=3.5cm, height=4cm]{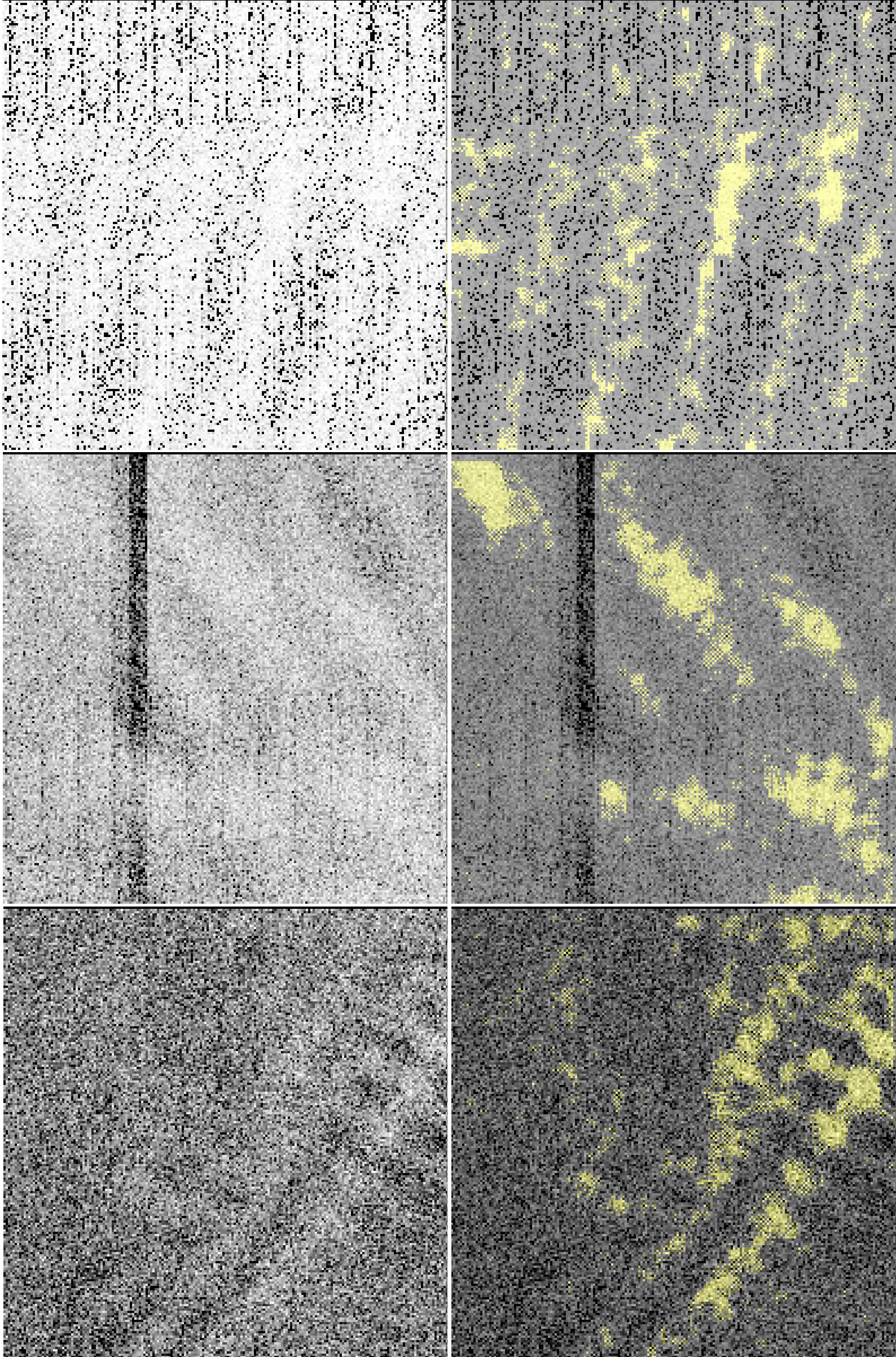}%
    \subcaption{complex detect}%
  \end{minipage}\hfil
  \begin{minipage}{.22\linewidth}
    \includegraphics[width=3.5cm, height=4cm]{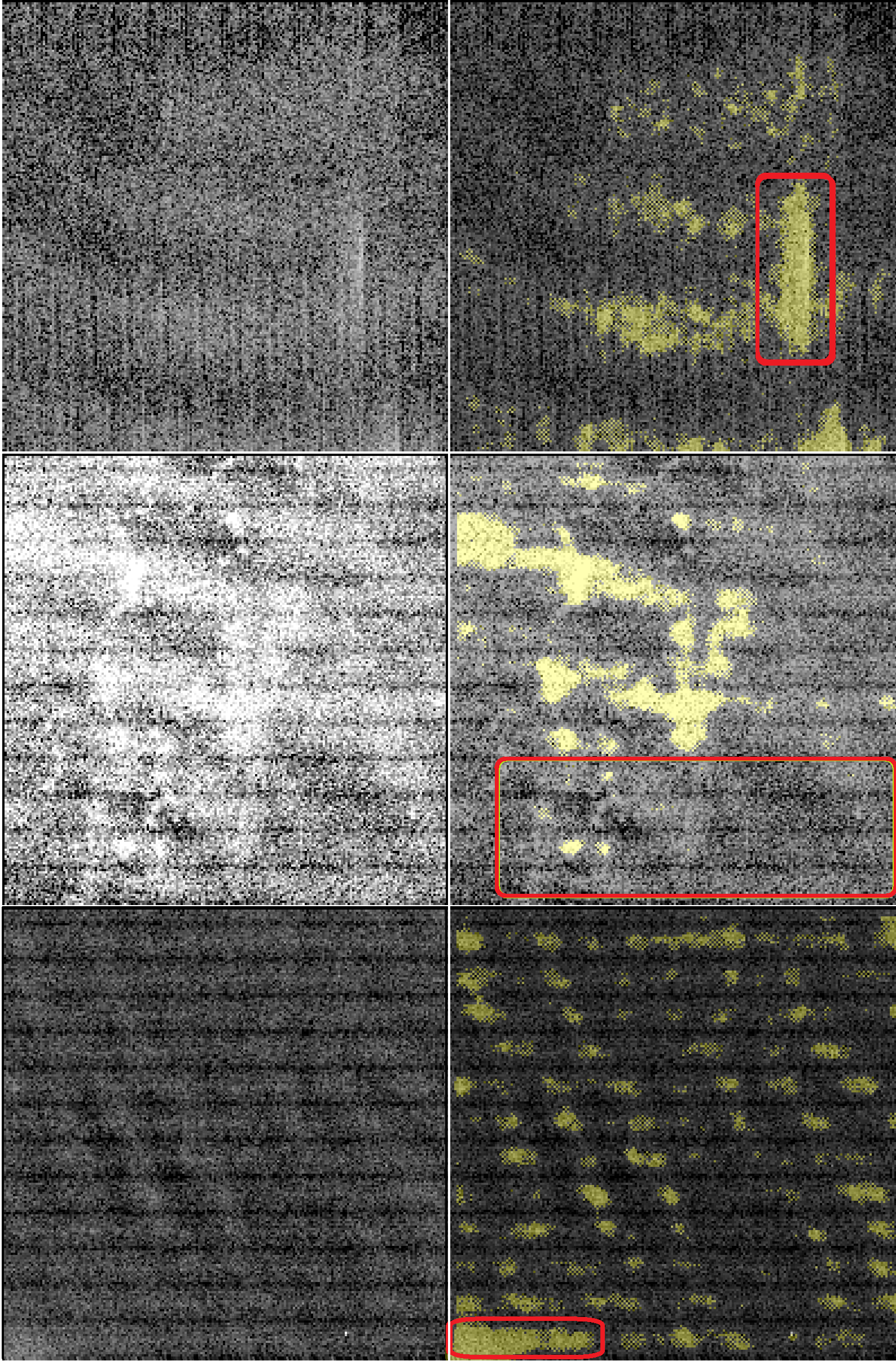}%
    \subcaption{partial fails}%
    \label{subfig:partialfail}
  \end{minipage}%
  \caption{The proposed kernel highlights extracted gravity wave features in sub-figures (a), (b), and (c). The left column of each sub-figure displays the actual PNG files, while the right column highlights the extracted gravity wave features. In sub-figure (a), simple patterns of gravity waves are detected. Sub-figure (b) shows detection of complex patterns even in the presence of noise, such as vertical lines in the top image and a black bar in the second image. However, sub-figure (c) exhibits partial failure in detection, as indicated by the red rectangular box.}
  \label{fig:kernelapplication}
  \vspace*{-.75cm} 
\end{figure*}

\textbf{Checkerboard Kernel for Gravity Wave Pattern Detection.} Since our dataset contains excessive noise, including city lights, clouds, and instrumental noise (horizontal/vertical lines), we designed the checkerboard kernel to capture all types of gravity wave patterns while excluding the noise. In this experiment, we utilized kernels of different sizes, including 3x3, 5x5, 7x7 and 9x9,  with the same pattern as illustrated in Figure~\ref{fig:ck}. We used the kernel in the first layer of our proposed deep-learning model (discussed later in this Section) to generate low-level features from input grayscale satellite images. The proposed custom kernel is defined as: $K(x, y)={\left[ (x+y+1)\%2 \right]}\_{x\in (0, ..., w) \: and \: y\in (0, ..., w)}$. Here, $w$ is the length of the square kernel (K) and $(x, y)$ is any position in the 2-dimensional kernel of size $(w \times w)$. The visualization of the proposed kernel is provided in Figure~\ref{fig:illust}. This kernel is capable of finding various shapes and orientations of the gravity wave traces from the noisy input dataset. We apply the checkerboard kernel on different images using the convolutional approach to observe the effects of the proposed kernel, as illustrated in Figure~\ref{fig:kernelapplication}. The figure demonstrates that the features corresponding to gravity waves are successfully extracted. To enhance visibility, we have highlighted these extracted features in yellow.

The concept behind utilizing a custom kernel within a deep learning approach is to enable the model to extract specific features relevant to the problem at hand. The conceptual purpose of the custom kernel, illustrated in Figure \ref{fig:illust}, is to capture intricate and nonlinear gravity wave patterns within the images that aligns the properties of gravity waves, even in the presence of noise. The alternating pattern of `1's and `0's within the checkerboard kernel enables the identification of lines, representing gravity waves, and gaps between them respectively. Additionally, the repeating pattern in the kernel helps identify recurring ripple-like nonlinear lines of varying shapes in the images. Our experimental results (in Section \ref{resutls}) indicate that the proposed \textit{gWaveNet} outperformed all other approaches.

\begin{figure}
  \centering
  \includegraphics[width=.9\linewidth]{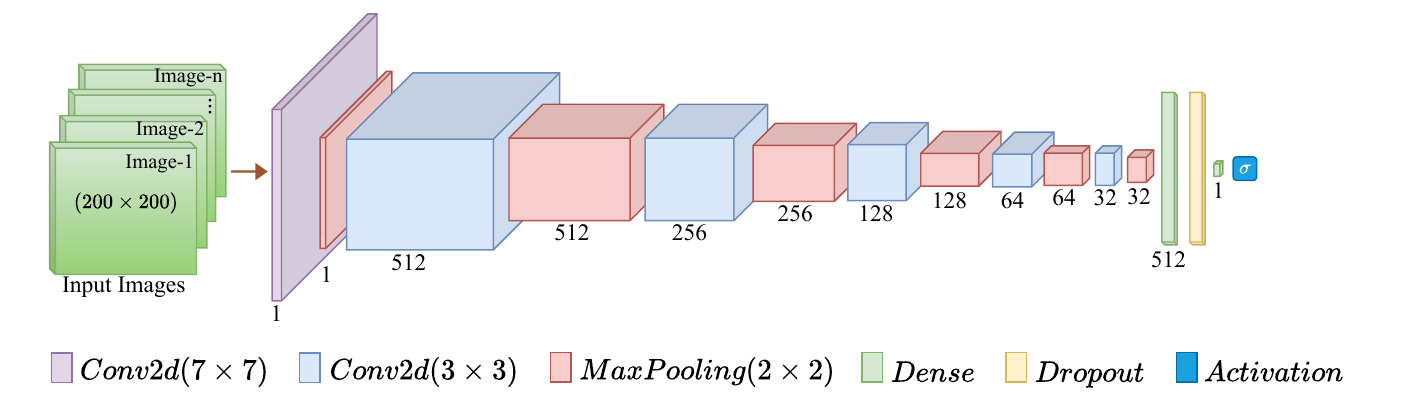}
  \caption{Architecture of the proposed \textit{gWaveNet} network.}
  \label{fig:model}
  \vspace*{-.5cm} 
\end{figure}

\textbf{Proposed Checkerboard Kernel-based Hybrid Neural Network.} Our proposed \textit{gWaveNet} in Figure \ref{fig:model} is a deep convolutional network with 15 layers. It features a hybrid model with the proposed checkerboard kernel integrated at the beginning, comprising 6 convolutional layers, followed by ReLU activations, 6 max-pooling layers, 2 dense layers, and 1 dropout layer, and a sigmoid activation at the end. Our custom kernel is placed in the first convolutional layer along with ReLU activation in the network.  
The hybrid architecture is designed for binary classification tasks on grayscale satellite images. The rationale behind incorporating the custom kernel in the initial layer is to specifically extract features aligned with those associated with gravity waves, allowing subsequent layers to identify similar intricate patterns. Focusing on the complexity of the problem, a large number of kernels is used in the earlier layers of the model which helps to learn more low-level features from the data. Gradually the number of kernels is reduced in the later layers to combine earlier features into more problem-specific high-level features. To avoid overfitting from the custom kernel, we apply L2 regularization in the second convolutional layer. 

We conducted experiments by configuring the layer in the proposed network to be either trainable or non-trainable, incorporating the checkerboard kernel (details are in Section \ref{resutls}). The checkerboard kernel's original weights of `0' or `1' can be updated during the model training process based on its configuration. Weights in the trainable layer are adjustable, directly influencing the model's predictions by learning from input data during training if the method is configured as trainable with the checkerboard kernel integrated. On the other hand, in the non-trainable configuration, although the weights are fixed, certain layers can still update their statistical entities such as mean and variance, indirectly impacting the model’s output, as noted in \cite{trainnotrain}. This indicates that non-trainable layers continue to impact the model's performance. We further explain, how the settings of the kernels with trainable/non-trainable would affect the overall model performance in Section \ref{resutls}.

\textbf{Experiment and Evaluation Setup.} For our experiment, all the implementation is carried out using Keras 2.11 and TensorFlow 2. The training and testing processes for both the proposed and baseline models were conducted on a GPU machine equipped with 20 gigabytes of memory.

We considered four distinct configurations to train all the models. In the ``trainable" configuration, we integrated the proposed checkerboard kernel into the first layer, with this layer set as trainable (\textit{trainable=true}). Conversely, the ``non-trainable" configuration retained the same concept, but with the first layer designated as non-trainable (\textit{trainable=false}). The ``no-custom-kernel-layer" (denoted as `nckl' in Tables) configuration indicated the absence of a custom kernel in the first layer. Lastly, the ``kernel-applied-prior-training" (denoted as `kapt' in Tables) configuration was similar to "no-custom-kernel-layer," except the kernel was applied to the images as a filter before initiating the model training process.

All models are trained for 2,000 epochs with a batch size of 128 on the training dataset. The training and validation dataset is split into a 65:35 ratio, and we reserved 240 image patches for testing, which were never used in training. The hyperparameters for the remaining methods are kept intact for experimentation with our data. We employ binary cross-entropy loss function during training to calculate the detection loss and The stochastic gradient descent (SGD) method, coupled with the binary cross-entropy loss objective function, is chosen for model optimization. In terms of evaluations, we used overall detection accuracy and the $F1$ score as evaluation metrics. 

\section{Experiment Results}
\label{resutls}
 \vspace{-.25cm}

In this section, we present our findings from various aspects such as , kernel size, trainable or non-trainable, kernel applied prior training (kapt) and no custom kernel layer (nckl) which is discussed in Ablation Study. We conducted extensive experiments to answer the following questions. Q1: \textit{How does the performance of our hybrid deep learning model with the integrated checkerboard kernel compare with state-of-the-art (SOTA) approaches?} Q2: \textit{How can the capability of the proposed model be inferred from the ablation studies?} Q3: \textit{How can the model learn without denoising the data?} Q4: \textit{Is the kernel integration approach generalizable?} Q5: \textit{How does the model perform when trained with a reduced amount of data?}

\textbf{Comparing with State-of-the-art Techniques.} We evaluated our model against five advanced State-of-the-Art (SOTA) techniques in computer vision research. This includes the Vision Transformer (ViT) \cite{dosovitskiy2020image}, an attention-based model, and VGG16 \cite{simonyan2015deep}, chosen for its similar architecture to our proposed model. Additionally, we assessed three influential convolutional filters: Gabor \cite{fogel1989gabor}, Sobel \cite{kanopoulos1988design}, and Laplacian \cite{wang2007laplacian}. Furthermore, our evaluation included an FFT denoising-based approach using Transfer Learning Mechanism \cite{gonzalez2022atmospheric}.
\vspace*{-.5cm} 

\begin{table*}
\caption{Performance comparison of the proposed model against state-of-the-art techniques: ViT \cite{dosovitskiy2020image},  VGG16 \cite{simonyan2015deep}, Gabor \cite{fogel1989gabor}, Sobel \cite{kanopoulos1988design}, Laplacian \cite{wang2007laplacian}, and Transfer Learning approach using FFT denoised data \cite{gonzalez2022atmospheric}.} 
    \label{tab:baselinecomp}
    \setlength\tabcolsep{4pt} 
    \begin{center}
    \scriptsize
    \begin{tabular}{|l|c|ccc|c|r|}
        \hline
        \multirow{2}{*}{\textbf{Methods}} & \multirow{2}{*}{\textbf{Architecture}} & \multicolumn{3}{c|}{\textbf{Accuracy}} & \multirow{2}{*}{\textbf{F1 Score}} & \multirow{2}{*}{\textbf{Train config.}} \\
        \cline{3-5}
        & & Train & Validation & Test & & \\
        \hline
        ViT & Transformer & \textbf{89.12} & 86.08 & 83.74 & 79.71 & nckl \\
        \hline
        VGG16 & VGG & \textbf{100.00} & \textbf{81.52} & 61.91 & 60.15 & nckl \\
        VGG16\_3x3\_t & VGG & 87.86 & \textbf{80.19} & \textbf{68.74} & \textbf{66.29} & trainable \\
        \hline
        Gabor\_7x7\_t & gWaveNet & \textbf{95.24} & \textbf{93.46} & \textbf{89.17} & \textbf{88.35} & trainable \\
        \hline
        Sobel\_3x3\_kapt & gWaveNet & 88.76 & 85.35 & 82.91 & 80.89 & kapt \\
        Sobel\_3x3\_t & gWaveNet & \textbf{91.23} & \textbf{87.17} & \textbf{83.74} & \textbf{82.24} & trainable \\
        \hline
        Laplacian\_7x7\_kapt & gWaveNet & 86.44 & 83.50 & 76.66 & 74.38 & kapt \\
        Laplacian\_7x7\_t & gWaveNet & \textbf{90.91} & \textbf{87.19} & \textbf{79.16} & \textbf{80.00} & trainable \\
        \hline
        FFT & gWaveNet & 76.59 & 75.04 & 70.66 & 69.50 & nckl \\
        FFT\_7x7\_nt & gWaveNet & 92.68 & 91.64 & 84.47 & 82.19 & non-trainable \\
        FFT\_7x7\_t & gWaveNet & \textbf{93.78} & \textbf{92.50} & \textbf{90.78} & \textbf{90.07} & trainable \\
        \hline
        gWaveNet\_5x5\_nt & gWaveNet & 94.40 & 93.16 & \textbf{93.75} & \underline{\textbf{91.89}} & non-trainable \\
        gWaveNet\_7x7\_t & gWaveNet & \underline{\textbf{98.10}} & \underline{\textbf{96.53}} & \underline{\textbf{94.21}} & \underline{\textbf{93.69}} & trainable \\
        \hline
    \end{tabular}
    \end{center}
    \vspace*{-.75cm} 
\end{table*}

We compared our model with State-of-the-Art techniques, summarized in Table~\ref{tab:baselinecomp} where the best results in each category are in bold and the overall best results are emboldened and underlined. First, we incorporated Gabor filters of size 7x7 with various orientations (0°, 30°, 60°, 120°, 150°) into our deep learning model considering the gravity wave patterns in the image. This process yielded high accuracy with an F1 score of 88.35\%.
Secondly, we applied Sobel filters (using 3x3 kernels) as a preprocessing step before training the model. Additionally, we integrated the Sobel filters into our proposed method during training, labeled as `Sobel\_3x3\_t' with a trainable kernel. The result indicates that the model trained with Sobel filters integrated into the first layer achieved higher accuracy and F1 score compared to the model trained on images pre-processed by Sobel filters (denoted as, Sobel\_3x3\_pt). However, our detailed optimization plot reports overfitting as the training keeps progressing. We followed the same process for the Laplacian filter, using a 7x7 kernel. The results showed a similar pattern, with better accuracy obtained when the Laplacian filter was integrated into the model (Laplacian\_7x7\_t) as opposed to utilizing it as a data preprocessing step (Laplacian\_7x7\_pt). However, the model also shows overfitting similar to Sobel filter. During our testing of the ViT model, the results indicated 89.12\% accuracy in training with 79.71\% F1 score which is quite comparable to the performance of Sobel and Laplacian approaches. However, the overfitting is significant.

Comparing the VGG16 method with its base architecture
and its modified architecture with our trainable approach (denoted as VGG16 and VGG16\_3x3\_t, respectively) reveals a notable difference. The base model is highly overfitted, showcasing high accuracy during training but experiencing a considerable drop in validation accuracy, nearly 20\%. 
On the contrary, VGG16 with a 3x3 kernel integrated trainable layer improved across all metrics though there are inconsistencies. While it still falls short of outperforming other models, the enhancement from the base model is notable. This suggests the potential generalizability of our proposed kernel with other methods addressing Q4. Furthermore, performance comparisons between VGG16\_3x3\_t and our gWaveNet\_noK model (Table \ref{tab:baselinecomp} and Table~\ref{tab:ablation}, respectively) show that VGG16\_3x3\_t trails behind gWaveNet\_noK in achieving competitive scores. Our experiments suggest that the gWaveNet\_noK model, specifically designed for our noisy dataset performs better in extracting relevant features from noisy data while the base VGG16 falls short due to its architectural configuration.

We further compared models trained with denoised data using FFT techniques based on three training configurations which are, trainable, non-trainable and no-custom-kernel-layer. The experiments revealed that the highest training accuracy achieved with FFT-denoised images reached 93.78\%. Notably, the model with a trainable layer outperformed the two other models with different training configurations for the same dataset. The results indicate a gradual improvement in model performance from the no-custom-kernel-layer to the non-trainable layer and finally to the trainable layer. However, with a 7x7 trainable layer the performance improved with consistent training. It is noteworthy that despite the high accuracy achieved by the model trained on denoised data, it still falls short compared to our proposed model trained using noisy data. This discrepancy could be attributed to the FFT transformation, which removes lower amplitude signals. There is a possibility that this process eliminated some gravity wave patterns that matched specific frequencies, leading to a lower overall performance score.

Referring to `gWaveNet\_7x7\_t', our approach outperforms all the aforementioned SOTA techniques. The proposed model, featuring a trainable 7x7 custom kernel, achieved the best results in terms of different accuracies and F1 score. Notably, these accuracies demonstrate improved optimization without overfitting. These findings address Q1 at the beginning of this section, highlighting the performance of our proposed hybrid deep learning model (in Table~\ref{tab:baselinecomp}) with the integrated checkerboard kernel.

\textbf{Ablation Study.} We thoroughly examined variety of configurations using the \textit{gWaveNet} core architecture in our ablation study, as shown in Table~\ref{tab:ablation}. In all training configurations, except for the `non-trainable' layers, we used various kernel sizes of 3x3, 5x5, 7x7, and 9x9, for all the \textit{gWaveNet} models. We discuss training configuration wise performances as follows.
\vspace*{-.75cm} 

\begin{table}
    \caption{Ablation Studies.}
    \label{tab:ablation}
    \setlength\tabcolsep{4pt} 
    \footnotesize
    \begin{center} 
    \begin{tabular}{|l|ccc|c|r|}
        \hline
        \multirow{2}{*}{\textbf{Methods}} & \multicolumn{3}{c|}{\textbf{Accuracy}} & \multirow{2}{*}{\textbf{F1 Score}} & \multirow{2}{*}{\textbf{Train config.}} \\
        \cline{2-4}
        & Train & Validation & Test & & \\
        \hline
        gWaveNet\_3x3\_t & 97.17 & 95.23 & 92.43 & 92.00 & trainable \\
        gWaveNet\_5x5\_t & \underline{\textbf{98.43}} & \underline{\textbf{97.22}} & 93.75 & 92.11 & trainable \\
        gWaveNet\_7x7\_t & 98.10 & 96.53 & \underline{\textbf{94.21}} & \underline{\textbf{93.69}} & trainable \\
        gWaveNet\_9x9\_t & 97.22 & 95.01 & 91.06 & 91.55 & trainable \\
        \hline 
        gWaveNet\_3x3\_nt & 91.38 & 90.25 & 87.50 & 86.63 & non-trainable \\
        gWaveNet\_5x5\_nt & 94.40 & 93.16 & \textbf{93.75} & \textbf{91.89} & non-trainable \\
        gWaveNet\_7x7\_nt & \textbf{97.53} & \textbf{95.60} & 92.50 & 91.61 & non-trainable \\
        gWaveNet\_9x9\_nt & 93.94 & 92.80 & 91.66 & 90.88 & non-trainable \\
        \hline 
        gWaveNet\_3x3\_pt & 97.21 & 95.69 & 90.83 & 89.47 & kapt \\
        gWaveNet\_5x5\_pt & 96.64 & 94.94 & 91.75 & 90.29 & kapt \\
        gWaveNet\_7x7\_pt & 97.72 & \textbf{96.03} & 93.24 & \textbf{93.07} & kapt \\
        gWaveNet\_9x9\_pt & \textbf{97.77} & 95.94 & \textbf{93.58} & 92.46 & kapt \\
        \hline 
        gWaveNet\_noK & \textbf{82.49} & \textbf{85.20} & \textbf{76.77} & \textbf{75.22} & nckl \\
        \hline 
        gWaveNet\_16K\_7x7 & \textbf{93.49} & \textbf{91.69} & \textbf{92.50} & \textbf{91.53} & trainable-16 \\
        gWaveNet\_64K\_7x7 & 93.33 & 91.61 & 90.13 & 89.07 & trainable-64 \\
        \hline
    \end{tabular}
    \end{center}
    \vspace{-.6cm}
\end{table}

\textit{Model with trainable or non-trainable layer.} The model with a trainable layer with integrated kernels refers to learning following kernel patterns, while the non-trainable layer refers to the opposite. With the trainable layer with the checkerboard kernel integrated, training, validation and testing accuracies are achieved as high as 98.43\%, 97.22\%, and 94.21\%, respectively, along with 93.69\% F1 score across models for various kernel sizes. The non-trainable layer also demonstrated significant performance with accuracies of 97.53\% in training, 95.60\% in validation, and 93.75\% in testing, along with an F1 score of 91.89\%. 

\textit{Models with kernel-applied-prior-training.} We also evaluated the proposed \textit{gWaveNet} model by applying the kernels to the images prior to training the models that also show significant performance. With this training approach, we achieved the best training accuracy of 97.77\% with a competitive F1 score of 93.07\%. 

\textit{Model with no-custom-kernel-layer.} As our next evaluation, we experimented with no-custom-kernel-layer (denoted as gWaveNet\_noK in Table~\ref{tab:ablation}). Without the custom kernel, the model exhibited relatively poor performance compared to other approaches in the table. The low F1 score indicates a large number of false positives and negatives in the confusion matrix possibly due to not learning the patterns of gravity waves, but the noise. This finding emphasizes the importance of using the proposed model architecture by integrating the checkerboard kernel for datasets dominated by noise. Though this approach performed poorly compared to other approaches that we proposed, it still performed better than VGG16 approaches in terms of learning and achieving better F1 scores.

\textit{Models with multiple kernels.} We further expanded our experiment to evaluate the proposed models with stacked identical 7x7 kernels, either 64 or 16 times. We observed that both approaches performed almost similarly, with a slight improvement using the kernel 16 times. The training accuracies were over 93\%, with F1 scores exceeding 91\%. From the experiment results, we notice that \textit{gWaveNet} model with the trainable custom kernel provided better performance than the models with the non-trainable custom kernel. This left us curious to investigate what changes are made to the custom kernel by the training process. For clarity, we derived the learned kernel, shown in  Figure~\ref{fig:trained_kernel}, at the first layer of the proposed trained model. If we compare this learned kernel with the proposed kernels (Figure~\ref{fig:ck}), the updated weights follow the same pattern as the initial kernel. This justifies that the custom kernel at the first layer has an impact on gaining better performance by the proposed model with trainable configuration, ensuring the continued relevance and applicability of the custom kernel.

\textit{In summary}, the ablation studies showed significant improvements in detecting intricate patterns with both trainable and non-trainable configurations. The method without a custom kernel ("gWaveNet\_noK") achieved an F1 score of 75.22, while integrating the custom kernel increased the F1 score to 91.89 (non-trainable) and 93.69 (trainable). Notably, the non-trainable configurations, apart from the trainable ones, outperformed all other state-of-the-art approaches in Table~\ref{tab:ablation}, demonstrating the custom kernel's effectiveness and improved generalizability over a standard convolutional kernel. Our findings from the ablation studies led us to conclude that the proposed model is highly effective in detecting gravity waves amidst noise, achieving higher accuracies. These conclusions address Q2 and Q3 that are related to the model's overall capability and ability to learn without denoising data.
\vspace*{-1cm} 

\begin{table}
    \caption{Comparison of models trained on 60\% and 40\% reduced datasets.}
    \label{tab:reduced_plot}
    \footnotesize
    \setlength\tabcolsep{4pt} 
    \begin{center} 
    \begin{tabular}{|l|ccc|c|}
        \hline
        \multirow{2}{*}{\textbf{Methods}} & \multicolumn{3}{c|}{\textbf{Accuracy}} & \multirow{2}{*}{\textbf{F1 Score}} \\
        \cline{2-4}
        & Train & Validation & Test & \\
        \hline
        gWaveNet\_9x9\_60\% & 89.74 & 87.85 & 86.66 & 86.39 \\
        gWaveNet\_7x7\_60\% & 94.18 & 93.24 & 89.99 & 88.67 \\
        gWaveNet\_5x5\_60\% & 93.99 & 92.46 & 86.66 & 84.39 \\
        gWaveNet\_3x3\_60\% & 94.02 & 92.27 & \textbf{93.75} & 90.76 \\
        gabor\_7x7\_60\%    & \textbf{99.86} & \textbf{93.44} & 91.04 & \textbf{93.50} \\
        sobel\_3x3\_60\%    & 89.70 & 82.66 & 68.15 & 69.37 \\
        vgg16\_3x3\_60\%    & 99.37 & 80.86 & 64.17 & 63.17 \\
        \hline
        gWaveNet\_9x9\_40\% & 99.50 & 85.14 & 63.74 & 61.35 \\    
        gWaveNet\_7x7\_40\% & 98.10 & 91.04 & 64.10 & 64.30 \\    
        gWaveNet\_5x5\_40\% & 99.29 & 83.33 & 69.08 & 70.36 \\    
        gWaveNet\_3x3\_40\% & \textbf{99.59} & 82.76 & 73.33 & 75.19 \\
        gabor\_7x7\_40\%    & 99.93 & \textbf{92.66} & \textbf{90.88} & \textbf{92.05} \\
        sobel\_3x3\_40\%    & 88.40 & 80.76 & 66.28 & 67.71 \\
        vgg16\_3x3\_40\%    & 100.00 & 77.99 & 74.76 & \textbf{-----} \\
        \hline
    \end{tabular}
    \end{center}
    \vspace*{-.5cm} 
\end{table}

\textbf{Model Evaluation with Reduced Amount of Data.} We further evaluated model's performance using a reduced dataset, which is 60\% and 40\% of the total data and the comparisons are depicted in Table~\ref{tab:reduced_plot}. The results in the Table highlights that \textit{gWaveNet} models trained with 60\% of the data exhibit minimal fluctuations in both training and validation accuracy, as well as in the F1 score. Conversely, other models with the same data proportion show significant overfitting, indicating suboptimal performance. When assessing models with 40\% of the data, all models, including \textit{gWaveNet}, exhibited overfitting issues. This portion of ablation study shows, training the model with more than 50\% of the data resulted in improved scores. In particular, VGG16 with a 3x3 trainable kernel failed to produce a meaningful F1 score due to its poor performance with the limited dataset. These findings addresses the Q5 regarding the reduced amount of data.
\vspace*{-.75cm} 

\begin{table}
 \footnotesize
  \caption{Comparison of mean and standard deviation between state-of-the-art and our proposed techniques.} \label{tab:meanstd}
  \begin{center} 
  \begin{tabular}{@{}|l|c|c|c|@{}}
    \toprule
    \textbf{Methods} & \textbf{Train\_acc} & \textbf{Validation\_acc} & \textbf{F1\_Score}\\
    \midrule
    VGG16\_3x3\_t & 86.98 $\pm$ 2.88 & 80.05 $\pm$ 1.59 & 67.66 $\pm$ \textbf{1.07} \\
    \hline
    Gabor\_7x7\_t & 91.83 $\pm$ 2.57 & 88.57 $\pm$ 3.50 & 76.29 $\pm$ 4.61 \\
    FFT\_7x7\_t & 92.78 $\pm$ \textbf{0.66} & 90.74 $\pm$ \textbf{0.61} & 89.56 $\pm$ \textbf{0.88}\\
    \hline 
    \textit{gWaveNet}\_3x3\_t & \textbf{98.25} $\pm$ 0.54 & \textbf{96.64} $\pm$ 0.71 & 91.58 $\pm$ 0.81\\
    \textit{gWaveNet}\_5x5\_t & 98.07 $\pm$ 0.18 & 96.31 $\pm$ 0.46 & 91.83 $\pm$ 0.99\\
    \textit{gWaveNet}\_7x7\_t & 97.71 $\pm$ 0.26 & 95.85 $\pm$ 0.54 & \textbf{92.95} $\pm$ \textbf{0.48}\\
    \textit{gWaveNet}\_9x9\_t & 97.10 $\pm$ \textbf{0.12} & 94.99 $\pm$ \textbf{0.09} & 90.80 $\pm$ 0.52\\
    \hline
    \textit{gWaveNet}\_64k\_7x7\_t & 92.61 $\pm$ 0.58 & 90.80 $\pm$ 0.76 & 88.89 $\pm$ 0.21\\
    \textit{gWaveNet}\_16k\_7x7\_t & 92.84 $\pm$ 0.46 & 90.53 $\pm$ 0.81 & 89.19 $\pm$ 0.60\\
    \hline
    \textit{gWaveNet}\_60\%\_3x3\_t & 93.89 $\pm$ \textbf{0.08} & 93.30 $\pm$ 1.61 & 88.11 $\pm$ 2.41\\
    \textit{gWaveNet}\_60\%\_5x5\_t & 93.82 $\pm$ 0.33 & 92.41 $\pm$ \textbf{0.37} & 87.74 $\pm$ 2.55\\ 
    \textit{gWaveNet}\_60\%\_7x7\_t & 95.57 $\pm$ 1.04 & 93.78 $\pm$ 0.42 & 89.62 $\pm$ 2.09\\
    \textit{gWaveNet}\_60\%\_9x9\_t & 91.53 $\pm$ 1.35 & 89.30 $\pm$ 1.16 & 87.61 $\pm$ \textbf{1.03}\\
    \bottomrule
  \end{tabular}
  \end{center}
  \vspace*{-.75cm} 
\end{table}

\textbf{Model Robustness Comparison.} As the final step, we compare the mean and standard deviation of selected methods in Table \ref{tab:meanstd}. Our comparison includes VGG16 with a 3x3 trainable kernel (denoted as VGG16\_3x3\_t), ViT method, Gabor approach and FFT-based methods along with all \textit{gWaveNet} methods with trainable layers. To ensure a thorough analysis of average performance and variability, each model was run five times.
As we see from the table, the deviations for VGG16\_3x3\_t, are not much, however, the model shows an inadequate performance compared to others in terms of F1 score. The Gabor approach performs well in both accuracies and F1 scores, but the standard deviation is higher across all cases. Comparing the model with FFT denoised data with a 7x7 trainable layer(denoted as FFT\_7x7\_t) exhibits higher accuracies with lower deviations. However, the F1 score does not achieve as high as our proposed model, even when using the custom kernel layer. Despite this, the deviation in all categories is better in FFT\_7x7\_t compared to the above models. When comparing the performance of our models, \textit{gWaveNet} with different kernel sizes, we observe minimal deviations, except few cases trained with 60\% data, indicating the robustness of our proposed hybrid deep learning model with the checkerboard kernel integrated.

\textbf{Limitations.} From our experiments, we observed a performance drop when \textit{gWaveNet} models are trained with less than 50\% of the data, emphasizing the common requirement of ample data in deep learning model training. Additionally, in some instances, applying the convolutional kernel detected partial patterns or missed certain patterns (sub-figure~\ref{subfig:partialfail}). We attribute this to the image preprocessing steps, indicating an area for improvement in future.
\vspace{1mm}

\textbf{Discussions.} Our proposed 7x7 and 5x5 kernels with trainable layers in \textit{gWaveNet} demonstrated significant performance improvements over baseline methods, including Gabor, Sobel, or Laplacian filter-based kernels, as well as advanced models like ViT and Vgg16. Despite achieving high training accuracy by fewer approaches like, the Gabor, VGG16 was hampered  with the overfitting issue or low F1 scores, indicating its limited effectiveness. However, \textit{gWaveNet} performed well in those cases, highlighting the critical role of both kernel shape and coefficients in enhancing model performance. Notably, our model showed superiority in non-trainable configurations, consistently achieving higher F1 scores compared to state-of-the-art techniques. These results highlight the effectiveness and improved generalizability of our custom kernel over standard convolutional kernels, establishing our approach as a new benchmark in gravity wave detection.

\section{Conclusions}
\label{conclusions}
Overall, our propose \textit{gWaveNet} model demonstrated the ability to learn without data denoising, with higher accuracies and the versatility of the proposed kernel allows for integration into other approaches towards its generalizability. However, in future, we would like to experiment with various satellite data of similar patterns collected from multi-angular view and also would like to address the underlying physics behind the patterns of the gravity waves and the proposed kernel. Additionally, we are interested in addressing the challenges associated with the localization of gravity waves using bounding boxes, particularly with a diverse set of images capturing similar patterns in presence of noise and unwanted objects.

\subsubsection{Acknowledgements:} This work is supported by the NASA grant “Machine Learning based Automatic Detection of Upper Atmosphere Gravity Waves from NASA Satellite Images” (80NSSC22K0641).

\bibliographystyle{splncs04}
\bibliography{egbib}

\end{document}